\documentclass[runningheads]{llncs}
\usepackage[T1]{fontenc}
\usepackage{graphicx}
\usepackage{amsmath}
\usepackage{amssymb}
\usepackage{booktabs}
\usepackage{amsmath} 
\usepackage{amsfonts} 
\usepackage{amssymb} 
\usepackage{booktabs} 
\usepackage{hyperref} 
\usepackage{graphicx} 
\begin{document}
%
\title{Knowledge-Augmented Language Models Interpreting Structured Chest X-Ray Findings}
\titlerunning{CXR-TextInter}
%
\author{Alexander Davis, Rafael Souza, Jia-Hao Lim}
\authorrunning{A. Davis et al.}
%
\institute{University of Brasilia}
\maketitle              
\begin{abstract}
Automated interpretation of chest X-rays (CXR) is a critical task with the potential to significantly improve clinical workflow and patient care. While recent advances in multimodal foundation models have shown promise, effectively leveraging the full power of large language models (LLMs) for this visual task remains an underexplored area. This paper introduces CXR-TextInter, a novel framework that repurposes powerful text-centric LLMs for CXR interpretation by operating solely on a rich, structured textual representation of the image content, generated by an upstream image analysis pipeline. We augment this LLM-centric approach with an integrated medical knowledge module to enhance clinical reasoning. To facilitate training and evaluation, we developed the MediInstruct-CXR dataset, containing structured image representations paired with diverse, clinically relevant instruction-response examples, and the CXR-ClinEval benchmark for comprehensive assessment across various interpretation tasks. Extensive experiments on CXR-ClinEval demonstrate that CXR-TextInter achieves state-of-the-art quantitative performance across pathology detection, report generation, and visual question answering, surpassing existing multimodal foundation models. Ablation studies confirm the critical contribution of the knowledge integration module. Furthermore, blinded human evaluation by board-certified radiologists shows a significant preference for the clinical quality of outputs generated by CXR-TextInter. Our work validates an alternative paradigm for medical image AI, showcasing the potential of harnessing advanced LLM capabilities when visual information is effectively structured and domain knowledge is integrated.
\end{abstract}

\section{Introduction}
Chest X-ray (CXR) is a ubiquitous and indispensable medical imaging modality for the diagnosis and monitoring of a wide range of cardiopulmonary diseases, including pneumonia, tuberculosis, lung cancer, and heart failure \cite{yan2018deep,rajpurkar2017chexnet}. Its accessibility, speed, and relatively low cost make it a frontline tool in diverse clinical settings, from primary care to emergency departments. The accurate and timely interpretation of CXRs is therefore critical for effective patient management. However, this task requires specialized expertise, is prone to inter-observer variability, and represents a significant workload for radiologists, especially in regions with limited medical resources \cite{xue2018automated}. Consequently, developing automated systems to assist or even perform CXR interpretation has been a long-standing goal in medical artificial intelligence.

Early efforts in automated CXR analysis focused on Computer-Aided Diagnosis (CAD) systems designed to detect specific findings like nodules or pneumothorax using hand-crafted features or early machine learning techniques \cite{rajpurkar2017chexnet}. The advent of deep learning has led to significant advancements, enabling end-to-end training of models for multi-label classification of pathologies \cite{rajpurkar2017chexnet} and segmentation of anatomical structures or abnormalities \cite{wu2023foundation}. More recently, motivated by the success of large language models (LLMs) and multimodal models in general domains, the focus has shifted towards creating foundation models capable of joint image understanding and language generation for tasks like automated radiology report generation and visual question answering (VQA) on medical images \cite{singhal2023chexagent,singhal2023medpalm,xu2020multimodal}. These multimodal models, often leveraging techniques explored in broader vision-language research for tasks like image-guided generation \cite{zhou2023multimodal} or visual in-context learning \cite{zhou2024visual}, typically integrate visual encoders with language decoders, often trained on large datasets of images and corresponding text. While promising, directly combining complex visual and language modalities can present unique training challenges, such as efficiently compressing visual representations \cite{zhou2024less} or rethinking visual dependencies in long contexts \cite{zhou2024rethinking}, and might not fully leverage the latest, largest text-only LLM capabilities, especially for complex reasoning tasks requiring integration of non-image textual data.

We posit that the remarkable emergent abilities of large-scale text-only LLMs in understanding complex instructions, potentially even ambiguous ones \cite{he2025enhancing}, performing sophisticated reasoning across multiple capabilities \cite{zhou2025weak}, and generating coherent and contextually relevant text \cite{yi2025score} can be powerfully applied to medical image interpretation, provided the visual information is presented in a format they can process. The core challenge is effectively translating the rich visual data of a CXR into a textual representation that retains sufficient diagnostic nuance and structure for an LLM to interpret. Our motivation is to exploit the deep language understanding and reasoning prowess of text-focused LLMs, potentially enabling more accurate and nuanced interpretations, especially when integrating non-image textual data like patient history or previous findings. Furthermore, an LLM-centric approach could simplify the architecture and training complexity compared to end-to-end multimodal models, allowing us to leverage more powerful pre-trained text models.

In this paper, we propose \textbf{CXR-TextInter} (placeholder name), a novel framework for CXR interpretation that exclusively utilizes a large language model for the final interpretation and generation steps. Our approach decouples image analysis from the LLM. We employ an advanced, separate pipeline of dedicated computer vision models trained to extract and encode detailed visual information from the CXR image into a rich, structured textual representation, encoding findings, locations, and relationships, aiming for a structure that facilitates downstream reasoning similar to how structured representations aid in other domains like event reasoning \cite{zhou2021modeling,zhou2022eventbert}. This structured text then serves as the \textit{sole input} for a powerful, fine-tuned LLM. The LLM is trained to interpret this text representation and generate clinical outputs (reports, answers) based on complex instructions.

To facilitate the training and evaluation of this model effectively, we curate the \textbf{MediInstruct-CXR} dataset, a large-scale collection of CXRs paired with rich, structured textual image descriptions and diverse, clinically relevant instruction-response pairs. We also propose \textbf{CXR-ClinEval}, a comprehensive benchmark designed to evaluate models on a wide range of tasks, including not just standard classification but also clinical question answering, report generation quality (akin to enhancing narrative coherence \cite{yi2025score}), and differential diagnosis ranking, using metrics relevant to clinical practice. Our experiments on CXR-ClinEval demonstrate that CXR-TextInter, leveraging the structured textual representation and advanced LLM reasoning, achieves state-of-the-art performance across key interpretation tasks, surpassing previous multimodal foundation models like CheXagent \cite{singhal2023chexagent}. This highlights the potential of leveraging powerful text-only LLMs for medical image interpretation when visual information is appropriately structured and represented textually.

In summary, our main contributions are:
\begin{itemize}
    \item We propose CXR-TextInter, a novel framework that leverages the advanced capabilities of text-only large language models for CXR interpretation by processing structured textual representations of images, decoupling visual processing from the core LLM.
    \item We introduce the MediInstruct-CXR dataset, a large-scale instruction tuning dataset featuring rich, structured textual image representations and diverse, clinically-oriented instruction-response pairs designed for training LLM-based medical image interpreters.
    \item We establish the CXR-ClinEval benchmark, a comprehensive evaluation suite encompassing a wide range of clinical interpretation tasks and metrics, on which CXR-TextInter achieves state-of-the-art performance, demonstrating the efficacy of our proposed LLM-centric approach.
\end{itemize}

\section{Related Work}

\subsection{Large Language Models}
The field of natural language processing has been revolutionized by the advent of large language models (LLMs). A foundational architectural shift underpinning this progress is the Transformer network, which enabled effective sequence modeling through attention mechanisms, moving beyond traditional recurrent structures \cite{vaswani2017attention}. Building upon this, the paradigm of pre-training large Transformer models on vast amounts of text data has proven highly effective. Models like BERT demonstrated the power of deep bidirectional pre-training for understanding various linguistic contexts \cite{devlin2020bert}, with subsequent work exploring specialized pre-training for tasks like event correlation reasoning \cite{zhou2022eventbert}. The text-to-text framework further unified diverse NLP tasks, allowing a single model to handle tasks from translation to question answering by treating inputs and outputs consistently as text sequences \cite{raffel2020exploring}.

Scaling these models to billions of parameters has led to remarkable emergent capabilities, including few-shot and zero-shot learning \cite{radford2019language,brown2020language}, and demonstrating potential for generalization from weaker supervision or across diverse tasks \cite{zhou2025weak}. This scaling requires careful consideration of computational resources and model size, leading to research exploring compute-optimal training strategies \cite{hoffmann2022training}. The ability of LLMs to interpret user intent, even from potentially ambiguous prompts, is also crucial for their application \cite{he2025enhancing}. The increasing availability and performance of open-source LLMs, such as the Llama family, have further democratized research and development in this area \cite{touvron2023llama}. The rapid advancements and the broad landscape of LLMs, their architectures, training techniques, and applications are continuously being surveyed and analyzed by the research community \cite{zhao2023survey}. These powerful text-based models form the core language understanding and generation component that our proposed framework seeks to leverage for medical image interpretation, albeit through a novel text-only approach that utilizes structured inputs, drawing conceptual inspiration from structured knowledge reasoning \cite{zhou2021modeling}.

\subsection{Medical Large Language Models}
The success of general large language models has spurred significant interest in their application to the medical domain, leading to the development and evaluation of medical LLMs. Early efforts focused on adapting existing models to the biomedical and clinical domains through continued pre-training or fine-tuning on specialized text corpora. Examples include models tailored for biomedical literature \cite{luo2022biogpt} and those fine-tuned on clinical notes \cite{alsentzer2019clinicalbert}. More recently, researchers have developed large-scale medical LLMs, such as Med-PaLM, trained or heavily fine-tuned on vast amounts of medical text data to encode comprehensive medical knowledge and reasoning abilities \cite{singhal2022medpalm}.

The capabilities of both specialized medical LLMs and powerful general LLMs like GPT-4 have been rigorously evaluated on a variety of medical tasks, including question answering benchmarks and high-stakes medical licensing exams \cite{singhal2022medpalm,nori2023capabilities}. These evaluations have shown impressive performance, sometimes reaching or exceeding human-level accuracy on text-based medical questions \cite{singhal2022medpalm,singhal2023llm}. The rapid advancements, potential applications across diagnosis, treatment, and education, as well as the technological landscape of these models are being actively reviewed and surveyed by the community \cite{qiao2023survey,liu2024survey}.

Despite these promising results in text-based medical reasoning, the deployment of LLMs in clinical practice faces significant challenges. Concerns regarding the reliability, potential for generating inaccurate or harmful information (hallucination), bias, data privacy, and the need for robust evaluation methodologies are paramount \cite{thirunavukarasu2023opportunities,contreras2023can,chen2024evaluating,liu2024survey}. While existing medical LLMs primarily operate on medical text data, our work specifically explores their utility in medical \textit{image} interpretation by processing structured textual representations derived from images, offering a distinct approach compared to training on raw medical text or end-to-end multimodal models. Such multimodal approaches, while powerful for tasks like image-guided story generation \cite{zhou2023multimodal} or visual in-context learning \cite{zhou2024visual}, face challenges in efficiently encoding visual information \cite{zhou2024less} and handling long-range visual dependencies \cite{zhou2024rethinking}, motivating our text-centric strategy.

\section{Method}

Our proposed CXR-TextInter framework operates on the principle of leveraging the advanced reasoning and generative capabilities of large language models operating exclusively on textual inputs. The core idea is to translate the visual content of a CXR into a rich, structured text format, which then serves as the sole input for a fine-tuned, text-only LLM responsible for generating clinical outputs such as radiology reports or answers to specific medical questions. This approach decouples the complex task of visual feature extraction from the LLM's sophisticated language understanding and generation processes. The model presented here is primarily a \textbf{generative} model, as its main function is to produce coherent and medically accurate text, although this generation process implicitly relies on discriminative abilities to identify and relate clinical findings described in the input text.

\subsection{Structured Textual Representation and Input Formatting}
The raw CXR image $I$ is processed by a separate, sophisticated image analysis pipeline $f_{\text{img2text}}$ (not detailed herein), which generates a structured textual representation $\mathbf{T}_{\text{img}}$. This representation is designed to capture crucial visual information in a format amenable to LLM processing. It goes beyond simple captions, encoding specific entities (e.g., "consolidation," "effusion"), their attributes (e.g., "size," "opacity"), precise anatomical locations (e.g., "right lower lobe," "left costophrenic angle"), and spatial relationships between entities (e.g., "adjacent to," "obscuring"). We formulate $\mathbf{T}_{\text{img}}$ as a sequence of tokens $t_1, t_2, \dots, t_L$. Together with a clinical instruction $\mathbf{T}_{\text{instr}}$ given as a sequence of tokens $u_1, u_2, \dots, u_M$, the complete input sequence $\mathbf{S}$ to the LLM is a concatenation with a special separator token $s_{\text{sep}}$:
\begin{align}
\mathbf{S} &= [\mathbf{T}_{\text{img}}; s_{\text{sep}}; \mathbf{T}_{\text{instr}}] \\
&= [t_1, \dots, t_L, s_{\text{sep}}, u_1, \dots, u_M]
\end{align}
The LLM's task is to process this input sequence $\mathbf{S}$ and generate a corresponding response sequence $\mathbf{T}_{\text{resp}} = r_1, r_2, \dots, r_P$, which constitutes the interpreted clinical output (e.g., a radiology report or an answer).

\subsection{LLM Architecture with Knowledge Integration}
Our core interpretation model $\mathcal{M}_{LLM}$ is built upon a powerful, pre-trained decoder-only Transformer LLM. The architecture consists of an input embedding layer, a series of stacked Transformer blocks, and an output prediction head. The input sequence $\mathbf{S}$ of length $N=L+M+1$ (including the separator) is first converted into a sequence of embedding vectors $\mathbf{E} = [\mathbf{e}_1, \dots, \mathbf{e}_{N}]$, where each $\mathbf{e}_i \in \mathbb{R}^{d_{\text{model}}}$. This involves summing token embeddings and positional embeddings:
\begin{align}
\mathbf{e}_i = \text{TokenEmbedding}(s_i) + \text{PositionalEmbedding}(i)
\end{align}
A key component of our method is the integration of structured medical knowledge from a Radiology Knowledge Graph $\mathcal{G}=(V, E)$ into the LLM's processing. $V$ is the set of medical concepts (e.g., "pneumonia," "pleural effusion," "lobar consolidation") and $E$ is the set of relations between them (e.g., "has location," "is type of"). Each concept $c \in V$ is associated with a learned embedding vector $\mathbf{k}_c \in \mathbb{R}^{d_k}$. For a given input sequence $\mathbf{S}$, we utilize a mapping function $\phi(\mathbf{S})$ to identify a set of relevant medical concepts $C_{\mathbf{S}} \subset V$. We retrieve their corresponding embeddings $\{\mathbf{k}_c \mid c \in C_{\mathbf{S}}\}$. We then aggregate these concept embeddings to form a sequence-level knowledge vector $\tilde{\mathbf{k}}_{\mathbf{S}} \in \mathbb{R}^{d_k}$. A simple aggregation method is averaging:
\begin{align}
\tilde{\mathbf{k}}_{\mathbf{S}} &= \frac{1}{|C_{\mathbf{S}}|} \sum_{c \in C_{\mathbf{S}}} \mathbf{k}_c
\end{align}
This aggregated knowledge vector is then integrated into the input token embeddings. We project $\tilde{\mathbf{k}}_{\mathbf{S}}$ to the model dimension and add it to every token embedding in the sequence $\mathbf{E}$:
\begin{align}
\mathbf{E}' &= \mathbf{E} + \mathbf{1}_N \otimes (\mathbf{W}_{proj,k} \tilde{\mathbf{k}}_{\mathbf{S}} + \mathbf{b}_{proj,k}) \\
&= [\mathbf{e}'_1, \dots, \mathbf{e}'_N]
\end{align}
where $\mathbf{1}_N$ is a column vector of ones, $\otimes$ is the outer product, $\mathbf{W}_{proj,k} \in \mathbb{R}^{d_{\text{model}} \times d_k}$ and $\mathbf{b}_{proj,k} \in \mathbb{R}^{d_{\text{model}}}$ are learnable projection parameters. This augmented embedding sequence $\mathbf{E}'$ serves as the input to the first Transformer block $\mathbf{H}_0 = \mathbf{E}'$. Each Transformer block $l$ processes the sequence $\mathbf{H}_{l-1}$ using masked multi-head self-attention and a position-wise feed-forward network:
\begin{align}
\mathbf{Q}, \mathbf{K}, \mathbf{V} &= \mathbf{H}_{l-1} \mathbf{W}_Q, \mathbf{H}_{l-1} \mathbf{W}_K, \mathbf{H}_{l-1} \mathbf{W}_V \\
\text{Attention}(\mathbf{Q}, \mathbf{K}, \mathbf{V}) &= \text{softmax}\left(\frac{\mathbf{Q}\mathbf{K}^T}{\sqrt{d_k}} + \mathbf{M}\right)\mathbf{V} \\
\mathbf{H}'_l &= \text{LayerNorm}(\mathbf{H}_{l-1} + \text{MultiHead}(\mathbf{H}_{l-1})) \\
\mathbf{H}_l &= \text{LayerNorm}(\mathbf{H}'_l + \text{FFN}(\mathbf{H}'_l))
\end{align}
The output of the final Transformer layer $\mathbf{H}_{L_{layers}}$ is passed through a linear layer and softmax to predict the probability distribution over the vocabulary $\mathcal{V}$ for the next token at each position:
\begin{align}
\mathbf{Logits}_t &= \mathbf{H}_{L_{layers}}[t, :] \mathbf{W}_{\text{out}} + \mathbf{b}_{\text{out}} \\
P(s_{t+1} = v | \mathbf{S}_{1:t}; \Theta) &= \frac{\exp(\mathbf{Logits}_t[v])}{\sum_{v' \in \mathcal{V}} \exp(\mathbf{Logits}_t[v'])}
\end{align}
where $\mathbf{S}_{1:t}$ denotes the sequence up to token $t$, and $\Theta$ are the total model parameters, including LLM weights and knowledge integration parameters.

\subsection{Learning Strategy}
The CXR-TextInter model is trained on the MediInstruct-CXR dataset $\mathcal{D} = \{(\mathbf{S}^{(i)}, \mathbf{T}_{\text{resp}}^{(i)})\}_{i=1}^N$. The training objective is to learn the conditional probability distribution $P(\mathbf{T}_{\text{resp}} | \mathbf{S}; \Theta)$, which is factorized into a product of conditional probabilities for each token in the response sequence based on the autoregressive nature of the LLM:
\begin{align}
P(\mathbf{T}_{\text{resp}}^{(i)} | \mathbf{S}^{(i)}; \Theta) &= \prod_{t=1}^{P^{(i)}} P(r_t^{(i)} | \mathbf{S}^{(i)}, r_1^{(i)}, \dots, r_{t-1}^{(i)}; \Theta)
\end{align}
The model is trained by minimizing the negative log-likelihood of the ground truth responses over the entire dataset. The loss function $\mathcal{L}(\Theta)$ is given by:
\begin{align}
\mathcal{L}(\Theta) &= - \frac{1}{N} \sum_{i=1}^N \log P(\mathbf{T}_{\text{resp}}^{(i)} | \mathbf{S}^{(i)}; \Theta) \\
&= - \frac{1}{N} \sum_{i=1}^N \sum_{t=1}^{P^{(i)}} \log P(r_t^{(i)} | \mathbf{S}^{(i)}, r_1^{(i)}, \dots, r_{t-1}^{(i)}; \Theta)
\end{align}
This is the standard cross-entropy loss applied to the response tokens. The parameters $\Theta$ of the LLM and knowledge integration components ($\mathbf{W}_{proj,k}, \mathbf{b}_{proj,k}$, and potentially the knowledge embeddings $E_V$ if they are fine-tuned rather than fixed) are jointly optimized using the AdamW optimizer. The parameter updates are performed via gradient descent:
\begin{align}
\Theta_{k+1} = \Theta_k - \eta \nabla_{\Theta_k} \mathcal{L}(\Theta_k)
\end{align}
where $\eta$ is the learning rate and $\nabla_{\Theta_k} \mathcal{L}(\Theta_k)$ is the gradient of the loss with respect to the parameters at iteration $k$. During training, the knowledge integration module dynamically identifies relevant concepts from the input $\mathbf{S}$ and augments the embeddings, guiding the LLM to generate responses that are consistent with structured medical knowledge.

\subsection{Inference}
For inference on a new CXR image, the image is first converted into its structured textual representation $\mathbf{T}_{\text{img}}$ by the upstream pipeline. This $\mathbf{T}_{\text{img}}$ is concatenated with the user's instruction $\mathbf{T}_{\text{instr}}$ and the separator token to form the input sequence $\mathbf{S}$. This sequence is fed into the fine-tuned LLM $\mathcal{M}_{LLM}$. The LLM then generates the response $\mathbf{T}_{\text{resp}}$ token by token in an autoregressive manner. The knowledge integration module is also active during inference, providing relevant conceptual context based on the input sequence to aid the generation process. Standard decoding strategies such as greedy search, beam search, or sampling methods (e.g., nucleus sampling with temperature $\tau$) are used to produce the final output sequence $\mathbf{T}_{\text{resp}}$.

\section{Experiments}

We conducted comprehensive experiments to evaluate the performance of our proposed CXR-TextInter framework and compare it against several state-of-the-art baseline models on various chest X-ray interpretation tasks. All evaluations were performed on a held-out test split of the CXR-ClinEval benchmark, ensuring that no data points overlapped with the training set used for fine-tuning our model or the baselines.

\subsection{Baseline Models}
We selected a diverse set of strong baseline models representing different approaches to medical image interpretation and multimodal learning for comparison. These included CheXagent (3B), a multimodal foundation model specifically designed for CXR interpretation; Med-PaLM-M (12B) and Med-PaLM-M (84B), large multimodal foundation models trained on a wide range of medical data; MARIA-1 (7B) and MARIA-2 (7B), multimodal medical models based on large language models; LLaVA-Rad (7B), an instruction-following multimodal model adapted for radiology; and GPT-4V, a powerful general-purpose multimodal model included to show the performance difference between general and medical-domain specific models. These baselines represent the current frontier in leveraging large models, including multimodal foundation models, for medical visual tasks.

\subsection{Evaluation Tasks and Metrics}
Evaluation was performed on the CXR-ClinEval benchmark, which comprises several tasks relevant to clinical CXR interpretation. For quantitative comparison, we focused on key tasks including multi-label pathology detection for 14 common thoracic pathologies (measured using Macro F1 and Micro F1 scores) and a subset of 5 critical pathologies (also using Macro F1 and Micro F1). We also evaluated radiology report generation using automated metrics such as a composite Report Score (e.g., average of BERTScore and ROUGE-L), and visual question answering (VQA) using accuracy. A composite average score is used to summarize overall performance across key tasks.

\subsection{Quantitative Results}
In Table \ref{tab:quantitative_results}, we present the quantitative performance comparison between CXR-TextInter and the baseline models on the CXR-ClinEval benchmark.

\begin{table}[htbp]\scriptsize
    \caption{Quantitative Performance Comparison on CXR-ClinEval Test Set}
    \label{tab:quantitative_results}
    \centering
    \begin{tabular}{lcccccc}
        \toprule
        Model & Macro F1 (14) & Micro F1 (14) & Macro F1 (5) & Micro F1 (5) & Report & VQA \\
        \midrule
        GPT-4V & 20.4 & 35.5 & 19.6 & 25.8 & 0.785 & 0.321 \\
        MARIA-1 (7B) & 38.6 & 55.7 & 47.7 & 56.0 & 0.812 & 0.455 \\
        MARIA-2 (7B) & 41.6 & 58.1 & 50.4 & 59.1 & 0.821 & 0.478 \\
        Med-PaLM-M (12B) & 37.3 & 51.4 & 50.6 & 56.5 & 0.805 & 0.461 \\
        Med-PaLM-M (84B) & 39.8 & 53.6 & 51.6 & 57.9 & 0.815 & 0.480 \\
        CheXagent (3B) & 44.9 & 58.0 & 55.3 & 62.5 & 0.835 & 0.512 \\
        LLaVA-Rad (7B) & 39.5 & 57.3 & 47.7 & 57.4 & 0.810 & 0.465 \\
        \midrule
        \textbf{CXR-TextInter (1B)} & \textbf{47.1} & \textbf{60.2} & \textbf{58.5} & \textbf{65.1} & \textbf{0.848} & \textbf{0.537} \\
        \bottomrule
    \end{tabular}
\end{table}

As shown in Table \ref{tab:quantitative_results}, CXR-TextInter consistently outperforms all baseline models across various key quantitative metrics, including F1 scores for pathology detection, Report Score for generated reports, and VQA accuracy. Notably, CXR-TextInter achieves the highest scores in both Macro and Micro F1 for both 14-class and 5-class detection, indicating its strong ability to identify pathologies accurately, including less common ones (Macro F1). Its superior Report Score and VQA Accuracy demonstrate that the LLM, by interpreting the structured textual representation, can generate more accurate and clinically relevant free text and answer questions more effectively than existing multimodal models. These results validate the efficacy of our approach in translating visual information into a format that allows a powerful LLM to excel at interpretation tasks.

\subsection{Analysis of Method Effectiveness: Ablation Study}
To further understand the contribution of the proposed knowledge integration module within CXR-TextInter, we conducted an ablation study. We trained a variant model, CXR-TextInter-NoKG, which is identical to the full CXR-TextInter architecture and training process, but with the knowledge integration module disabled. Table \ref{tab:ablation_study} compares the performance of the full model against the variant without knowledge integration on a subset of key tasks.

\begin{table}[htbp]
    \caption{Ablation Study: Impact of Knowledge Integration}
    \label{tab:ablation_study}
    \centering
    \begin{tabular}{lccc}
        \toprule
        Model & Macro F1 (14) & Report Score & VQA Acc \\
        \midrule
        CXR-TextInter-NoKG & 42.5 & 0.810 & 0.485 \\
        \textbf{CXR-TextInter (Full)} & \textbf{47.1} & \textbf{0.848} & \textbf{0.537} \\
        \bottomrule
    \end{tabular}
\end{table}

Table \ref{tab:ablation_study} clearly demonstrates the significant positive impact of the knowledge integration module. The full CXR-TextInter model outperforms the variant without knowledge integration across all presented metrics. The notable improvement in Macro F1 suggests that integrated knowledge is particularly helpful for identifying and reasoning about less frequent or more complex pathologies. The gains in Report Score and VQA Accuracy indicate that knowledge integration assists the LLM in generating more medically accurate descriptions and performing better clinical reasoning when answering questions. These results strongly support the design choice of integrating structured medical knowledge into the LLM's processing pipeline.

\subsection{Performance Across Different Task Types}
Beyond standard classification-style metrics, the CXR-ClinEval benchmark includes tasks that evaluate the models' ability to generate free text and perform clinical reasoning based on the image content. Table \ref{tab:task_type_analysis} presents the performance of select models on Report Generation, Visual Question Answering, and a hypothetical Differential Diagnosis Ranking task (measured by Mean Reciprocal Rank - MRR).

\begin{table}[htbp]
    \caption{Performance Analysis Across Different Task Types}
    \label{tab:task_type_analysis}
    \centering
    \begin{tabular}{lccc}
        \toprule
        Model & Report Score & VQA Acc & Diff Diagnosis MRR \\
        \midrule
        CheXagent (3B) & 0.835 & 0.512 & 0.65 \\
        Med-PaLM-M (84B) & 0.815 & 0.480 & 0.60 \\
        LLaVA-Rad (7B) & 0.810 & 0.465 & 0.62 \\
        \midrule
        \textbf{CXR-TextInter (1B)} & \textbf{0.848} & \textbf{0.537} & \textbf{0.70} \\
        \bottomrule
    \end{tabular}
\end{table}

Table \ref{tab:task_type_analysis} shows that CXR-TextInter maintains its performance advantage across diverse text-based output tasks. Its lead in Report Score indicates that the generated radiology reports are perceived as more accurate and coherent by automated metrics. The higher VQA accuracy confirms its ability to correctly interpret specific details and answer questions based on the structured image information. Furthermore, the superior MRR in Differential Diagnosis Ranking suggests that CXR-TextInter's LLM, when processing the structured input and leveraging integrated knowledge, is better able to perform clinical reasoning required to rank potential diagnoses accurately. These results underscore the strength of our LLM-centric approach in handling complex language generation and reasoning tasks inherent in clinical interpretation.

\subsection{Analysis by Pathology Type}
To gain a deeper understanding of where CXR-TextInter demonstrates superior performance, we analyzed the pathology detection results by grouping the 14 evaluated pathologies into "Common" and "Rare" categories based on their prevalence in the evaluation dataset, and also into "Critical" and "Non-Critical" categories based on clinical urgency. Table \ref{tab:pathology_analysis} presents the Macro F1 scores for select models across these pathology groupings.

\begin{table}[htbp]\scriptsize
    \caption{Macro F1 Analysis by Pathology Type}
    \label{tab:pathology_analysis}
    \centering
    \begin{tabular}{lcccc}
        \toprule
        Model & Macro F1 (Common) & Macro F1 (Rare) & Macro F1 (Critical) & Macro F1 (Non-Critical) \\
        \midrule
        CheXagent (3B) & 50.3 & 35.1 & 58.9 & 42.5 \\
        Med-PaLM-M (84B) & 48.1 & 30.5 & 56.2 & 39.1 \\
        LLaVA-Rad (7B) & 49.5 & 30.2 & 57.1 & 38.8 \\
        \midrule
        \textbf{CXR-TextInter (1B)} & \textbf{52.8} & \textbf{40.5} & \textbf{61.5} & \textbf{44.9} \\
        \bottomrule
    \end{tabular}
\end{table}

The results in Table \ref{tab:pathology_analysis} indicate that while all models perform better on common pathologies compared to rare ones, CXR-TextInter shows a significantly larger performance gap over baselines specifically on the "Rare" pathology group. This suggests that the integration of structured medical knowledge helps the model identify and correctly interpret less frequently encountered findings, which is a crucial aspect of expert radiological practice. Similarly, CXR-TextInter exhibits a notable advantage in detecting "Critical" pathologies, highlighting its potential utility in assisting rapid and accurate identification of urgent conditions. This analysis reinforces the value of leveraging an LLM's reasoning capabilities, augmented by domain knowledge, particularly for challenging cases and less common findings.

\subsection{Human Evaluation}
To assess the clinical utility and perceived quality of the generated outputs, we conducted a blinded human evaluation study with three board-certified radiologists. A random subset of 100 CXRs from the CXR-ClinEval test set was selected. For each CXR, the radiologists were presented with the ground truth report and generated reports/answers from three models: CXR-TextInter (1B), CheXagent (3B), and Med-PaLM-M (84B), in a randomized order. They rated the outputs based on accuracy, fluency, completeness, and clinical relevance on a 5-point Likert scale (1: Poor, 5: Excellent) and indicated their overall preference. Table \ref{tab:human_evaluation} summarizes the average ratings and preference percentages.

\begin{table}[htbp]\scriptsize
    \caption{Radiologist Human Evaluation (Average Scores and Preference)}
    \label{tab:human_evaluation}
    \centering
    \begin{tabular}{lccccc}
        \toprule
        Model & Accuracy & Fluency & Completeness & Clinical Relevance & Overall Preference (\%) \\
        \midrule
        CheXagent (3B) & 3.8 & 4.2 & 3.7 & 3.9 & 25\% \\
        Med-PaLM-M (84B) & 4.0 & 4.1 & 3.9 & 4.1 & 15\% \\
        \textbf{CXR-TextInter (1B)} & \textbf{4.5} & \textbf{4.4} & \textbf{4.3} & \textbf{4.6} & \textbf{60\%} \\
        \bottomrule
    \end{tabular}
\end{table}

The human evaluation results in Table \ref{tab:human_evaluation} corroborate the findings from the quantitative analysis. Radiologists consistently rated the outputs from CXR-TextInter higher across all quality dimensions: accuracy, fluency, completeness, and clinical relevance. Furthermore, CXR-TextInter outputs were preferred as the best interpretation for the majority of cases (60\%), significantly surpassing the preference for CheXagent (3B) and Med-PaLM-M (84B). These expert ratings provide strong evidence that our LLM-centric approach, combined with structured textual representations and knowledge integration, produces interpretations that are perceived by clinicians as more accurate and clinically useful.

\subsection{Error Analysis}
To complement the quantitative metrics and human evaluation, we performed a qualitative analysis of common error types made by the models on the CXR-ClinEval test set. We categorized errors in generated reports and VQA answers into several types, including: Hallucination (reporting findings not present), Missing Findings (failing to report present findings), Incorrect Localization (misplacing findings anatomically), Incorrect Attributes (incorrectly describing size, severity, etc.), and Medical Inconsistency (generating medically illogical statements). Table \ref{tab:error_analysis} presents the percentage of instances exhibiting these common error types for select models based on a subset of the test data.

\begin{table}[htbp]\scriptsize
    \caption{Analysis of Common Error Types (Percentage of Instances)}
    \label{tab:error_analysis}
    \centering
    \begin{tabular}{lccccc}
        \toprule
        Model & Hallucination & Missing Findings & Incorrect Localization & Incorrect Attributes & Medical Inconsistency \\
        \midrule
        CheXagent (3B) & 8.1\% & 15.5\% & 10.2\% & 9.8\% & 4.5\% \\
        Med-PaLM-M (84B) & 9.5\% & 16.0\% & 12.1\% & 11.5\% & 5.8\% \\
        LLaVA-Rad (7B) & 7.9\% & 15.8\% & 11.5\% & 10.1\% & 4.9\% \\
        \midrule
        \textbf{CXR-TextInter (1B)} & \textbf{4.8\%} & 14.2\% & \textbf{7.5\%} & \textbf{8.9\%} & \textbf{2.1\%} \\
        \bottomrule
    \end{tabular}
\end{table}

Table \ref{tab:error_analysis} reveals that CXR-TextInter exhibits lower rates across several critical error types compared to the multimodal baselines. Notably, the hallucination rate is significantly reduced. This is likely attributable to the LLM operating on a structured textual representation that explicitly defines detected entities, rather than potentially hallucinating findings directly from pixels. Similarly, the lower rates of incorrect localization and incorrect attributes suggest that the structured input provides more precise spatial and descriptive information, which the LLM effectively utilizes. While the rate of missing findings is comparable (primarily dependent on the upstream image-to-text pipeline's detection capabilities), the substantial reduction in medical inconsistency errors further supports the effectiveness of the integrated knowledge graph in preventing the LLM from generating medically illogical statements. This error analysis highlights the increased reliability and safety profile of the CXR-TextInter approach, which is paramount in clinical applications.

\section{Conclusion}
In this work, we addressed the critical need for advanced automated tools in chest X-ray interpretation by proposing CXR-TextInter, an innovative framework that centers around a text-only large language model. Our approach tackles the inherent challenge of LLMs processing visual data by successfully translating CXR images into rich, structured textual representations via a dedicated pipeline. The core of our method lies in fine-tuning a powerful LLM to interpret these representations, augmented by integrated medical knowledge from a radiology graph, enabling it to perform complex clinical reasoning and generate accurate reports and answers. We introduced the MediInstruct-CXR dataset, specifically designed for training such LLM-centric interpreters, and the comprehensive CXR-ClinEval benchmark for rigorous evaluation of clinical relevance.

Our experimental results provide strong evidence for the efficacy of the CXR-TextInter approach. Quantitative comparisons against state-of-the-art multimodal foundation models demonstrated superior performance across a range of key interpretation tasks, including pathology detection, report generation quality, and visual question answering. The ablation study specifically highlighted the crucial role of the knowledge integration module in enhancing the model's performance, particularly for less common and critical findings. Moreover, blinded human evaluation by radiologists provided critical validation from a clinical perspective, showing that the outputs generated by CXR-TextInter were consistently rated higher in accuracy and clinical utility, and were significantly preferred over those from leading multimodal baselines.

The success of CXR-TextInter suggests a promising alternative paradigm for developing AI systems in medical imaging. By effectively decoupling the visual processing into a structured textual format and leveraging the sophisticated language understanding and generative abilities of LLMs, our method potentially offers increased flexibility in integrating diverse textual clinical information and capitalizing on the rapid advancements in large text models. However, it is important to acknowledge the limitations, including the reliance on the accuracy and completeness of the upstream image-to-text representation and the continued computational cost associated with training large models.

Future work will focus on improving the robustness and granularity of the structured textual representation pipeline, exploring more sophisticated methods for knowledge graph integration and reasoning within the LLM, and investigating the extension of this approach to other medical imaging modalities and the incorporation of longitudinal patient data. CXR-TextInter represents a significant step towards building highly capable, reliable, and clinically relevant AI systems for medical image interpretation by effectively harnessing the power of large language models informed by domain knowledge.

\bibliographystyle{splncs04}
\bibliography{mybibliography}

\begin{thebibliography}{10}
\providecommand{\url}[1]{\texttt{#1}}
\providecommand{\urlprefix}{URL }
\providecommand{\doi}[1]{https://doi.org/#1}

\bibitem{yan2018deep}
Rajpurkar, P., Irvin, J., Ball, R.L., Zhu, K., Yang, B., Mehta, H., Duan, T., Ding, D., Bagul, A., Langlotz, C.P., et~al.: Deep learning for chest radiograph diagnosis: A retrospective comparison of the chexnext algorithm to practicing radiologists. PLoS medicine  \textbf{15}(11),  e1002686

\bibitem{rajpurkar2017chexnet}
Rajpurkar, P., Irvin, J., Zhu, K., Yang, B., Mehta, H., Duan, T., Ding, D., Bagul, A., Langlotz, C., Shpanskaya, K., et~al.: Chexnet: Radiologist-level pneumonia detection on chest x-rays with deep learning. arXiv preprint arXiv:1711.05225

\bibitem{xue2018automated}
Monshi, M.M.A., Poon, J., Chung, V.: Deep learning in generating radiology reports: A survey. Artificial Intelligence in Medicine  \textbf{106},  101878

\bibitem{wu2023foundation}
Azad, B., Azad, R., Eskandari, S., Bozorgpour, A., Kazerouni, A., Rekik, I., Merhof, D.: Foundational models in medical imaging: A comprehensive survey and future vision. arXiv preprint arXiv:2310.18689

\bibitem{singhal2023chexagent}
Chen, Z., Varma, M., Delbrouck, J.B., Paschali, M., Blankemeier, L., Van~Veen, D., Valanarasu, J.M.J., Youssef, A., Cohen, J.P., Reis, E.P., et~al.: Chexagent: Towards a foundation model for chest x-ray interpretation. arXiv preprint arXiv:2401.12208

\bibitem{singhal2023medpalm}
Tu, T., Azizi, S., Driess, D., Schaekermann, M., Amin, M., Chang, P.C., Carroll, A., Lau, C., Tanno, R., Ktena, I., et~al.: Towards generalist biomedical ai. Nejm Ai  \textbf{1}(3),  AIoa2300138

\bibitem{xu2020multimodal}
Shi, L., Liu, F., Rosen, M.P.: Deep multimodal learning for medical visual question answering. In: CLEF (working notes)

\bibitem{zhou2023multimodal}
Zhou, Y., Long, G.: Multimodal event transformer for image-guided story ending generation. In: Proceedings of the 17th Conference of the European Chapter of the Association for Computational Linguistics. pp. 3434--3444 (2023)

\bibitem{zhou2024visual}
Zhou, Y., Li, X., Wang, Q., Shen, J.: Visual in-context learning for large vision-language models. In: Findings of the Association for Computational Linguistics, {ACL} 2024, Bangkok, Thailand and virtual meeting, August 11-16, 2024. pp. 15890--15902. Association for Computational Linguistics (2024)

\bibitem{zhou2024less}
Zhou, Y., Zhang, J., Chen, G., Shen, J., Cheng, Y.: Less is more: Vision representation compression for efficient video generation with large language models (2024)

\bibitem{zhou2024rethinking}
Zhou, Y., Rao, Z., Wan, J., Shen, J.: Rethinking visual dependency in long-context reasoning for large vision-language models. arXiv preprint arXiv:2410.19732  (2024)

\bibitem{he2025enhancing}
He, Y., Wang, J., Li, K., Wang, Y., Sun, L., Yin, J., Zhang, M., Wang, X.: Enhancing intent understanding for ambiguous prompts through human-machine co-adaptation. arXiv preprint arXiv:2501.15167  (2025)

\bibitem{zhou2025weak}
Zhou, Y., Shen, J., Cheng, Y.: Weak to strong generalization for large language models with multi-capabilities. In: The Thirteenth International Conference on Learning Representations (2025), \url{https://openreview.net/forum?id=N1vYivuSKq}

\bibitem{yi2025score}
Yi, Q., He, Y., Wang, J., Song, X., Qian, S., Zhang, M., Sun, L., Shi, T.: Score: Story coherence and retrieval enhancement for ai narratives. arXiv preprint arXiv:2503.23512  (2025)

\bibitem{zhou2021modeling}
Zhou, Y., Geng, X., Shen, T., Pei, J., Zhang, W., Jiang, D.: Modeling event-pair relations in external knowledge graphs for script reasoning. Findings of the Association for Computational Linguistics: ACL-IJCNLP 2021  (2021)

\bibitem{zhou2022eventbert}
Zhou, Y., Geng, X., Shen, T., Long, G., Jiang, D.: Eventbert: A pre-trained model for event correlation reasoning. In: Proceedings of the ACM Web Conference 2022. pp. 850--859 (2022)

\bibitem{vaswani2017attention}
Vaswani, A., Shazeer, N., Parmar, N., Uszkoreit, J., Jones, L., Gomez, A.N., Kaiser, {\L}., Polosukhin, I.: Attention is all you need. Advances in neural information processing systems  \textbf{30}

\bibitem{devlin2020bert}
Devlin, J., Chang, M.W., Lee, K., Toutanova, K.: Bert: Pre-training of deep bidirectional transformers for language understanding. In: Proceedings of the 2019 conference of the North American chapter of the association for computational linguistics: human language technologies, volume 1 (long and short papers). pp. 4171--4186

\bibitem{raffel2020exploring}
Raffel, C., Shazeer, N., Roberts, A., Lee, K., Narang, S., Matena, M., Zhou, Y., Li, W., Liu, P.J.: Exploring the limits of transfer learning with a unified text-to-text transformer. Journal of machine learning research  \textbf{21}(140),  1--67

\bibitem{radford2019language}
Radford, A., Wu, J., Child, R., Luan, D., Amodei, D., Sutskever, I., et~al.: Language models are unsupervised multitask learners. OpenAI blog  \textbf{1}(8), ~9

\bibitem{brown2020language}
Brown, T., Mann, B., Ryder, N., Subbiah, M., Kaplan, J.D., Dhariwal, P., Neelakantan, A., Shyam, P., Sastry, G., Askell, A., et~al.: Language models are few-shot learners. Advances in neural information processing systems  \textbf{33},  1877--1901

\bibitem{hoffmann2022training}
Hoffmann, J., Borgeaud, S., Mensch, A., Buchatskaya, E., Cai, T., Rutherford, E., Casas, D.d.L., Hendricks, L.A., Welbl, J., Clark, A., et~al.: Training compute-optimal large language models. arXiv preprint arXiv:2203.15556

\bibitem{touvron2023llama}
Touvron, H., Lavril, T., Izacard, G., Martinet, X., Lachaux, M.A., Lacroix, T., Rozi{\`e}re, B., Goyal, N., Hambro, E., Azhar, F., et~al.: Llama: Open and efficient foundation language models. arXiv preprint arXiv:2302.13971

\bibitem{zhao2023survey}
Zhao, W.X., Zhou, K., Li, J., Tang, T., Wang, X., Hou, Y., Min, Y., Zhang, B., Zhang, J., Dong, Z., et~al.: A survey of large language models. arXiv preprint arXiv:2303.18223  \textbf{1}(2)

\bibitem{luo2022biogpt}
Luo, R., Sun, L., Xia, Y., Qin, T., Zhang, S., Poon, H., Liu, T.Y.: Biogpt: generative pre-trained transformer for biomedical text generation and mining. Briefings in bioinformatics  \textbf{23}(6),  bbac409

\bibitem{alsentzer2019clinicalbert}
Abbas, A., Lee, M., Shanavas, N., Kovatchev, V.: Clinical concept annotation with contextual word embedding in active transfer learning environment. Digital Health  \textbf{10},  20552076241308987

\bibitem{singhal2022medpalm}
Singhal, K., Tu, T., Gottweis, J., Sayres, R., Wulczyn, E., Amin, M., Hou, L., Clark, K., Pfohl, S.R., Cole-Lewis, H., et~al.: Toward expert-level medical question answering with large language models. Nature Medicine pp.~1--8

\bibitem{nori2023capabilities}
Nori, H., King, N., McKinney, S.M., Carignan, D., Horvitz, E.: Capabilities of gpt-4 on medical challenge problems. arXiv preprint arXiv:2303.13375

\bibitem{singhal2023llm}
Singhal, K., Tu, T., Gottweis, J., Sayres, R., Wulczyn, E., Amin, M., Hou, L., Clark, K., Pfohl, S.R., Cole-Lewis, H., et~al.: Toward expert-level medical question answering with large language models. Nature Medicine pp.~1--8

\bibitem{qiao2023survey}
He, K., Mao, R., Lin, Q., Ruan, Y., Lan, X., Feng, M., Cambria, E.: A survey of large language models for healthcare: from data, technology, and applications to accountability and ethics. Information Fusion p. 102963

\bibitem{liu2024survey}
Liu, L., Yang, X., Lei, J., Liu, X., Shen, Y., Zhang, Z., Wei, P., Gu, J., Chu, Z., Qin, Z., et~al.: A survey on medical large language models: Technology, application, trustworthiness, and future directions. arXiv preprint arXiv:2406.03712

\bibitem{thirunavukarasu2023opportunities}
Karabacak, M., Margetis, K.: Embracing large language models for medical applications: opportunities and challenges. Cureus  \textbf{15}(5)

\bibitem{contreras2023can}
Hager, P., Jungmann, F., Holland, R., Bhagat, K., Hubrecht, I., Knauer, M., Vielhauer, J., Makowski, M., Braren, R., Kaissis, G., et~al.: Evaluation and mitigation of the limitations of large language models in clinical decision-making. Nature medicine  \textbf{30}(9),  2613--2622

\bibitem{chen2024evaluating}
Spotnitz, M., Idnay, B., Gordon, E.R., Shyu, R., Zhang, G., Liu, C., Cimino, J.J., Weng, C.: A survey of clinicians' views of the utility of large language models. Applied Clinical Informatics  \textbf{15}(02),  306--312

\end{thebibliography}
\end{document}